\documentclass{article}
\usepackage{stywhispers}
\usepackage{amsmath,epsfig}

\usepackage{siunitx}
\usepackage[hidelinks]{hyperref}
\usepackage{tabularx}
\usepackage{booktabs}
\usepackage{subfig}
\usepackage{multirow}
\usepackage{color}
\usepackage{paralist}
\usepackage[super]{nth}
\usepackage[nameinlink, capitalize]{cleveref}
\usepackage{float}

\title{Fusion of hyperspectral and ground penetrating radar data to estimate soil moisture}
\name{Felix M. Riese, Sina Keller}
\address{Karlsruhe~Institute~of~Technology~(KIT)\\
	Institute~of~Photogrammetry~and~Remote~Sensing~(IPF),\\
	Englerstra{\ss}e 7, D-76131 Karlsruhe, Germany
	}

\begin{document}
%
%
\maketitle
\noindent\copyright\ 2018 IEEE
\begin{abstract} 
In this contribution, we investigate the potential of hyperspectral data combined with either simulated ground penetrating radar (GPR) or simulated (sensor-like) soil-moisture  data to estimate soil moisture.
We propose two simulation approaches to extend a given multi-sensor dataset which contains sparse GPR data.
In the first approach, simulated GPR data is generated either by an interpolation along the time axis or by a machine learning model. 
The second approach includes the simulation of soil-moisture along the GPR profile.
The soil-moisture estimation is improved significantly by the fusion of hyperspectral and GPR data.
In contrast, the combination of simulated, sensor-like soil-moisture values and hyperspectral data achieves the worst regression performance.
In conclusion, the estimation of soil moisture with hyperspectral and GPR data engages further investigations.
\end{abstract}
\begin{keywords}
Hyperspectral data, ground penetrating radar, soil moisture, machine learning, regression, simulation
\end{keywords}
%

\section{Introduction}
\label{sec:intro}

Soil surface is a starting point when investigating infiltration behavior of precipitation~\cite{brooks2015hydrological}.
In this context, knowledge about the soil-moisture dynamics is crucial to understand ecological, hydrological, and vegetation processes. 
Traditionally, soil moisture is measured in-situ with high temporal resolution.
Time domain reflectometry (TDR) sensors represent such measuring devices.
Since it is extremely time-consuming and  expensive to cover large areas with these devices, remote sensing techniques like hyperspectral sensors are deployed, even partly on drones~\cite{finn2011remote, stamenkovic2013estimation}. 

To estimate soil-moisture with hyperspectral data, machine learning models are applied due to their potential to handle high-dimensional regression problems (cf.~\cite{stamenkovic2013estimation, riese2018introducing}).
These models are suitable to link hyperspectral image data as input data to soil-moisture data as target variable.

In addition to hyperspectral remote sensing, ground penetrating radar (GPR) measurements provide further data to model soil-moisture variations as described in~\cite{huisman2003measuring, allroggen20154d, xinbo2017measurement}.
In contrast to e.g. TDR probes, a GPR is feasible to consider the spatial variation of soil moisture~\cite{huisman2003measuring,jackisch2017form}.
Its measurements provide a more detailed spatial coverage and resolution.
The GPR data is manually acquired which is highly elaborative and therefore results in sparse datasets. 

Clearly, hyperspectral remote sensing and GPR measurements, have their particular strengths in different scopes of estimating soil moisture.
Our objective is to investigate the fusion of both techniques to address soil-moisture estimation.
We take advantage of the area-wide coverage of hyperspectral snapshot and the spatially high-resolution information of subsurface variations obtained from GPR data.

In the proposed studies, we rely on a dataset which is introduced in~\cite{keller2018modeling}.
The GPR data acquisition is performed in a frequency that is \num{10} times lower than the TDR-based soil-moisture measurements.
As a consequence, data gaps occur and it is necessary to artificially extend the GPR data.
Solely with this extension, a comprehensive soil-moisture estimation based on machine learning models can be conducted (cf. ~\cite{keller2018modeling, keller2018is}).

In this contribution, we present two novel approaches which simulate GPR and TDR data to extend the given, sparse dataset. 
Hyperspectral data combined with this simulated data is used to estimate soil moisture based on machine learning techniques.
The two approaches are entirely different. 
In the first approach, we address the simulation of GPR data along the time axis by using TDR and hyperspectral data.
The results are temporal-simulated GPR data which are combined with hyperspectral data to estimate the target variable soil moisture.
In contrast, the second approach contains the simulation of TDR measurements along the GPR measurement area.
Based on the spatial resolution of the GPR measurement, the simulated TDR data is projected onto this area and is used as reference to estimate soil moisture solely with hyperspectral data.
We then investigate the potential of each approach with respect to the performance of the soil-moisture estimation.

In \cref{sec:data}, we briefly introduce the dataset on which the proposed approaches are based.
These approaches are described in \cref{sec:methods}. 
We evaluate the performance of the regression task in \cref{sec:eval}.
In \cref{sec:conclusion}, we conclude our \mbox{studies}, respond to the underlying objective, and give an overview about future applications.

\section{Field campaign dataset}
\label{sec:data}

We use a dataset acquired during a multi-sensor field campaign in August 2017 in Linkenheim-Hochstetten, Germany~\cite{keller2018modeling}.
An undisturbed grassland site on loamy sand is divided into several plots. 
Each plot has a size of $\SI{1}{\meter}\times\SI{1}{\meter}$. 
They are monitored by a variety of sensors.
The plots are manually irrigated with a predefined irrigation schema~\cite{keller2018modeling}.
A hyperspectral snapshot camera performs the monitoring of the soil surface. 
In addition, a GPR acquires profile measurements.
TDR probes measure hydrological parameters in various soil depths.
For this publication, the TDR data in a depth of \SI{5}{\centi\meter} represents the reference values.

The used hyperspectral snapshot camera is a Cubert\footnote{Cubert GmbH, Ulm, Germany} UHD 285.
It captures images consisting of $50\times 50$ pixels with \num{125} spectral bands from \SIrange{450}{950}{\nano\meter} and a spectral resolution of \SI{4}{\nano\meter}.
As one data preprocessing step, we dismiss the first five and the last five spectral bands of the hyperspectral bands to reduce the effect of sensor artifacts.

The GPR system pulseEKKO PRO GPR\footnote{Sensors \& Software Inc., Mississaugua, Canada} equipped with a pair of 1 GHz shielded antennas collects information about the subsurface heterogeneity.
It is moved along fixed axes on the measurement field at predefined times, described in detail in~\cite{keller2018modeling}.
For each GPR measurement, soil moisture variations $\Delta\theta$ are derived with a procedure described in~\cite{allroggen20154d} resulting in a spatial resolution of \SI{1}{\centi\meter} along the measurement line.
Finally, we receive GPR profiles with \num{100} $\Delta\theta$ values per measurement plot.

\begin{figure}[tb]
	\centering
	\includegraphics[width=0.40\textwidth]{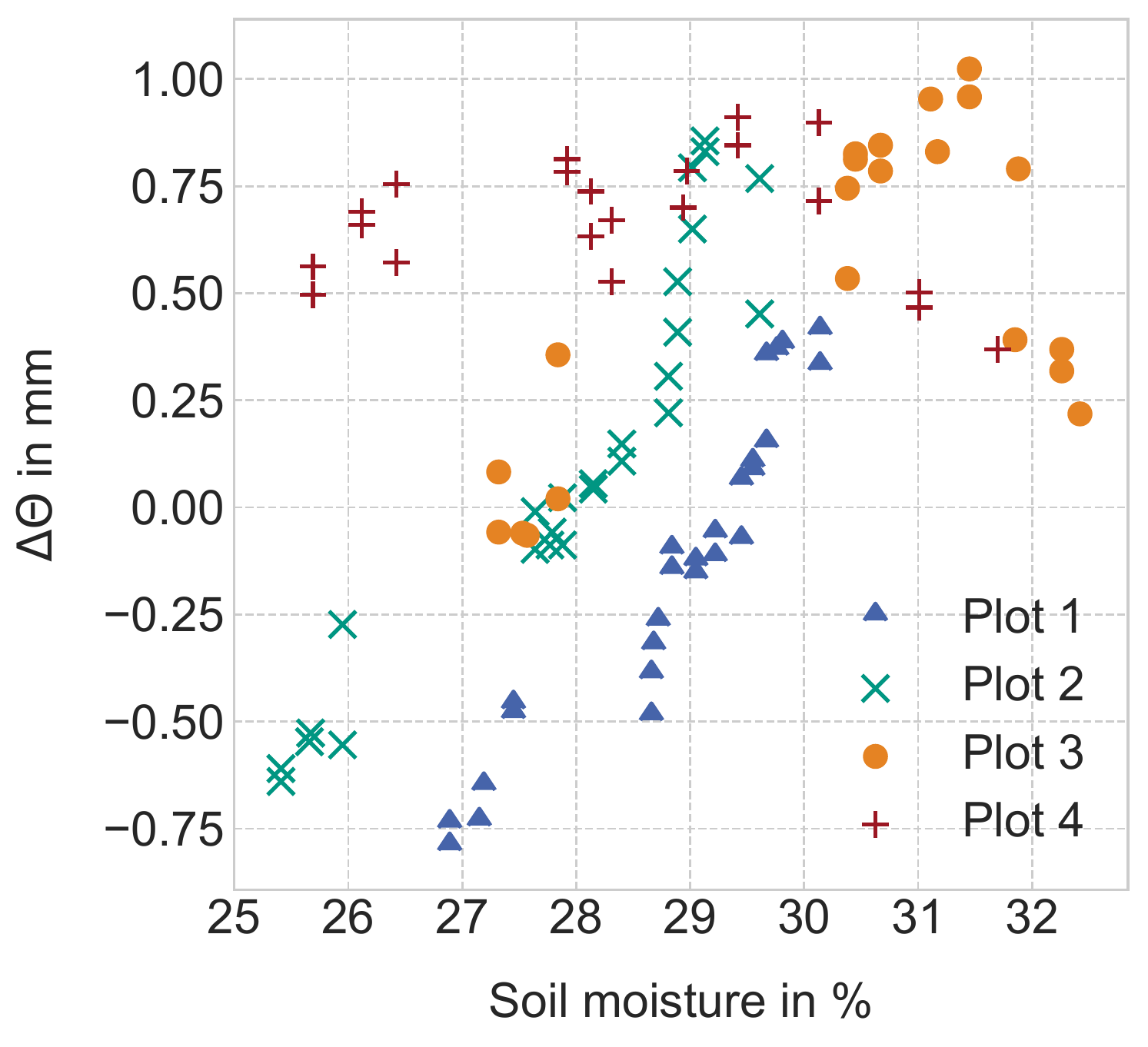}
	\caption{Correlations between the soil-moisture variations $\Delta\theta$ measured by the GPR and the soil-moisture values measured by the TDR sensors of the regarded four different measurement plots.\label{fig:corr}}
\end{figure}

We merge the spatial resolution of the GPR data to the \SI{10}{\centi\meter} resolution of the hyperspectral data.
Thus, the resulting dataset includes a hyperspectral measurement for every GPR profile.
In our studies, four of eight measurement plots are included.
The dataset then consists of \num{94} datapoints with one measured GPR $\Delta\theta$ value and one TDR soil-moisture value.
\cref{fig:corr} shows the correlation between the soil moisture variations $\Delta\theta$ of the GPR and the soil-moisture values of the TDR.
The different markers symbolize the four plots where the data is acquired.
The Pearson correlation coefficient of all datapoints is $r_{\text{all}}=\SI{50}{\percent}$.
\cref{tab:corr} summarizes this coefficient for each of the four plots.
This behavior can be explained by the different irrigation schemes for the measurement plots (cf.~\cite{keller2018modeling}).

\begin{table}[tb]
	\centering
	\caption{Pearson correlation coefficient $r$ of the four plots.}
	\begin{tabular}{lccccc}
	\toprule
	Plot & 1 & 2 & 3 & 4 & all\\ 
	\midrule
	$r$ in \% & 94 & 93 & 64 & -6 & 50\\
	\bottomrule
	\end{tabular}
	\label{tab:corr}	
\end{table}

\section{Simulation approaches}
\label{sec:methods}

\begin{figure*}[tb]
	\centering
	\subfloat[]{\includegraphics[height=6.0cm, keepaspectratio]{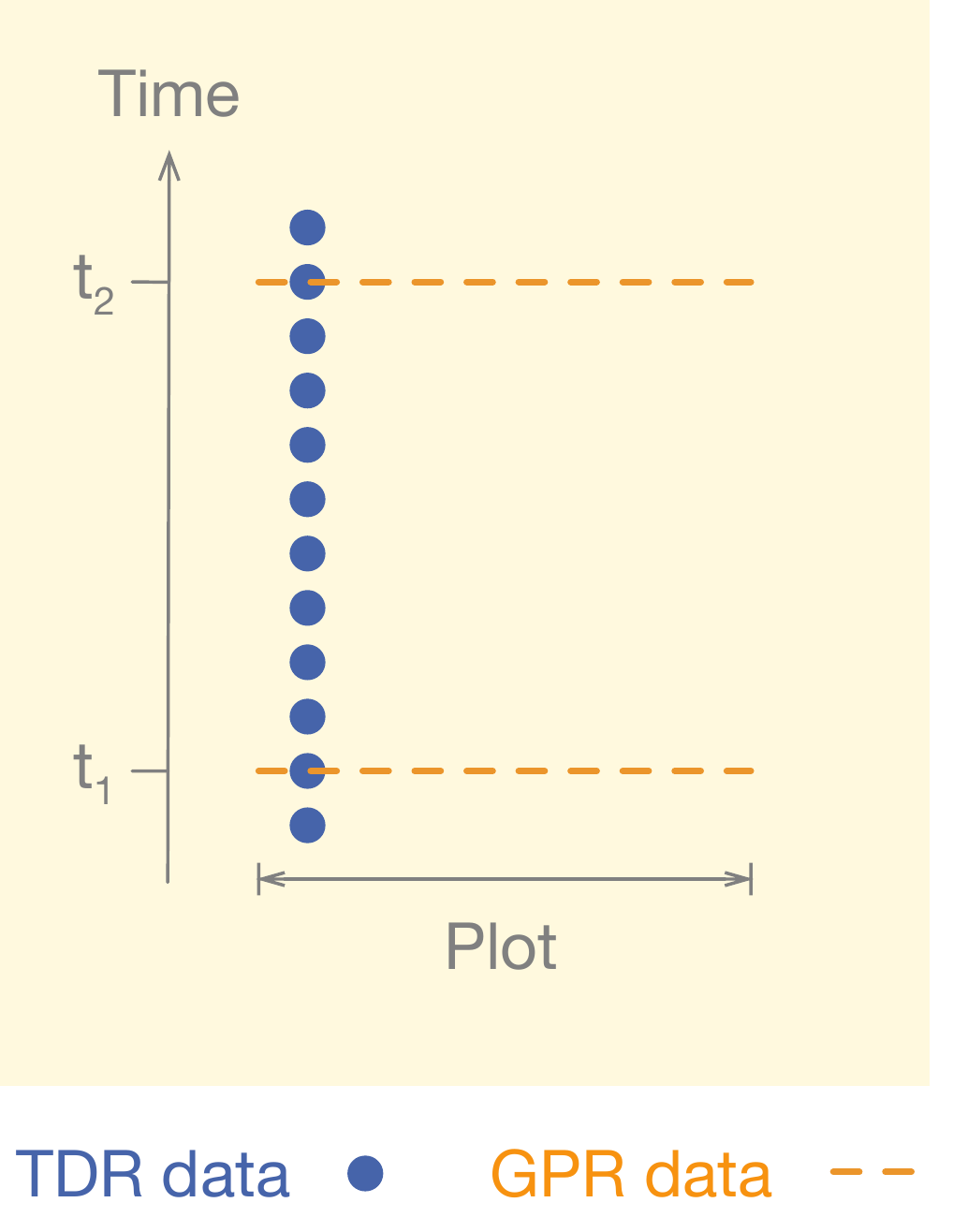}\label{fig:gpr_schema_meas}}\qquad
	\subfloat[]{\includegraphics[height=6.0cm, keepaspectratio]{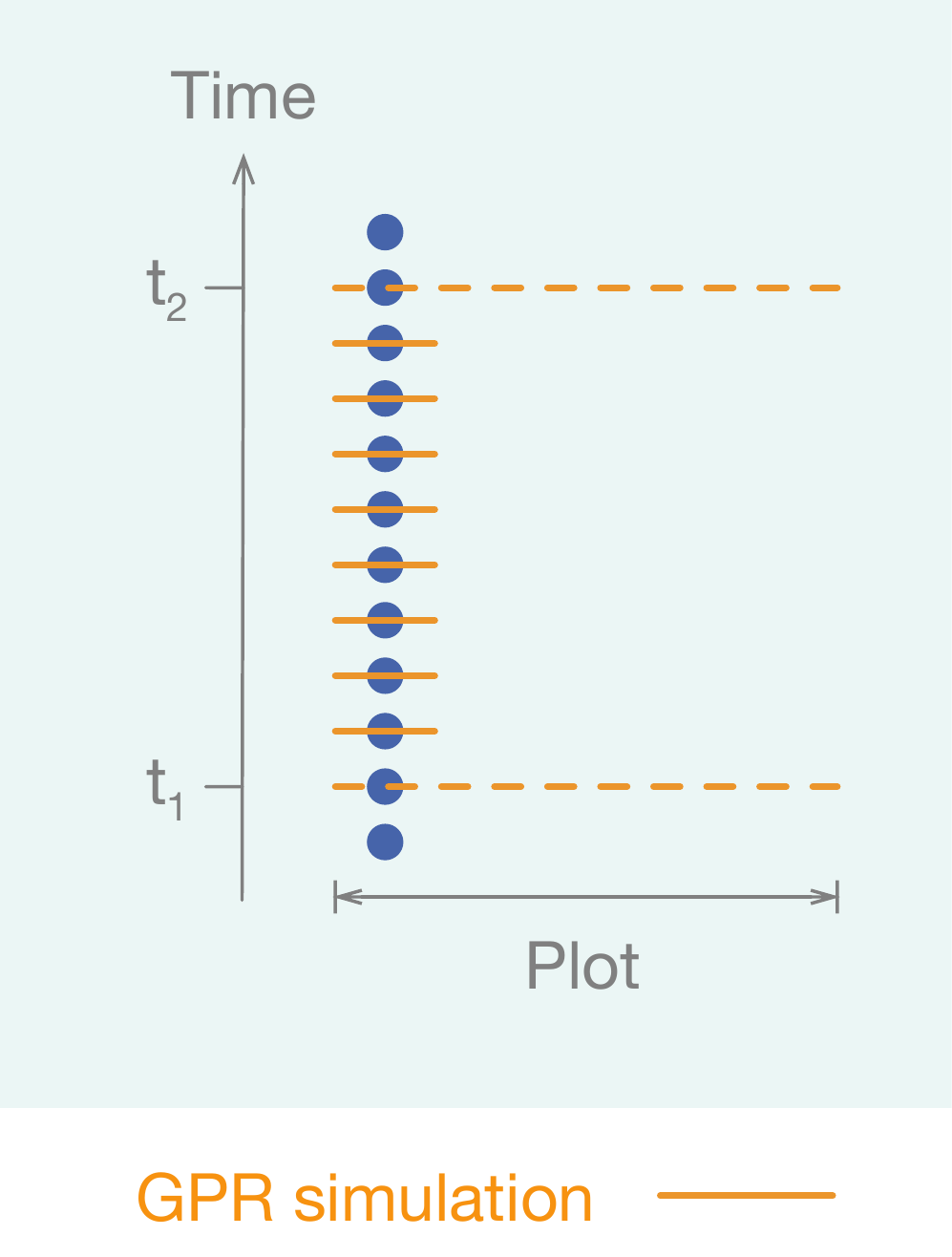}\label{fig:gpr_schema_app1}}\qquad
	\subfloat[]{\includegraphics[height=6.0cm, keepaspectratio]{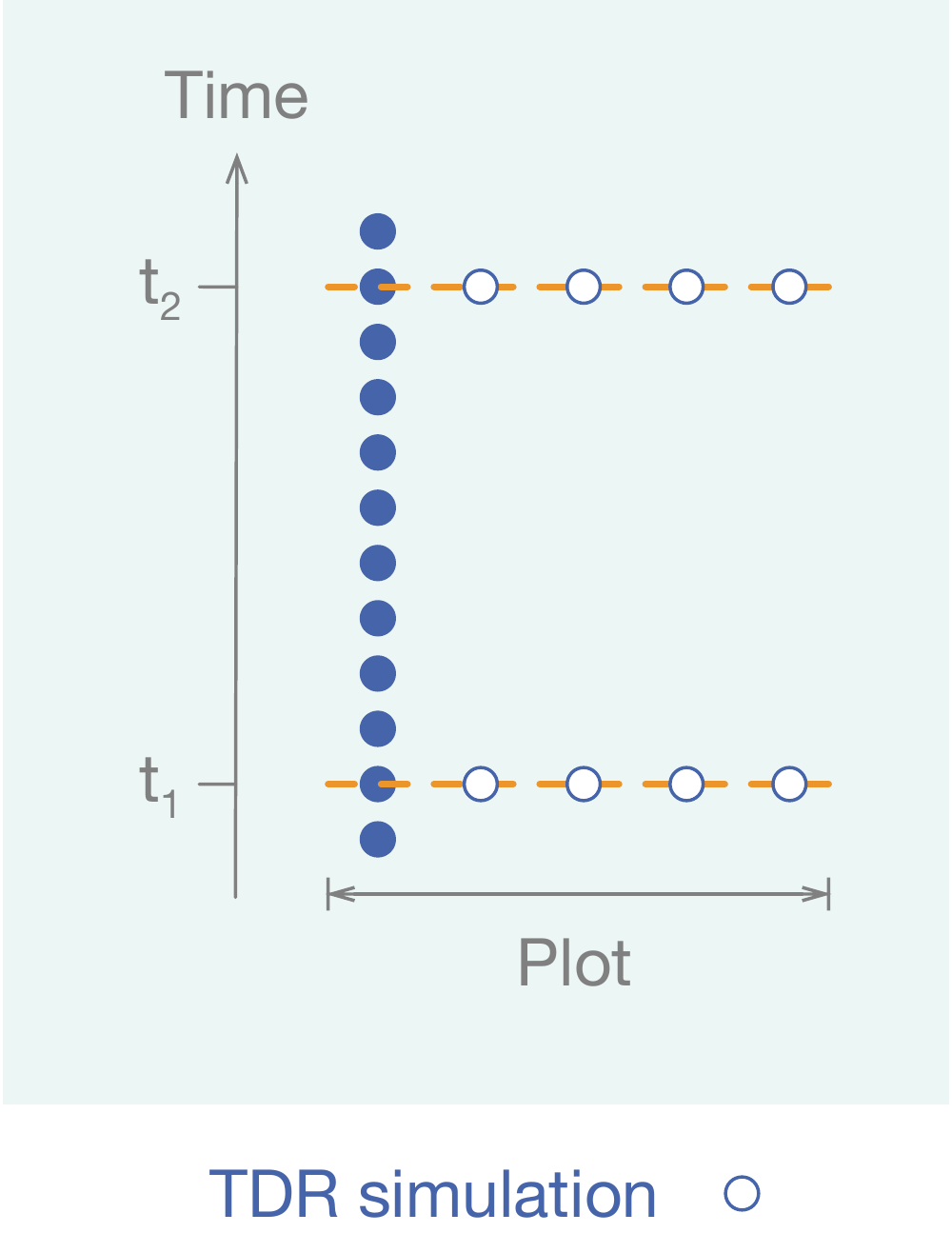}\label{fig:gpr_schema_app2}}
	\caption{Schematic overview of (a) the real measurements, (b) the GPR simulation as first approach, and (c) the TDR simulation as second approach. The vertical axes show the time with two exemplary points in time $t_1, t_2$ at which the GPR measurements (orange dashed lines) are performed. The horizontal axes show the \SI{1}{\meter} width of a measurement plot. The soil-moisture measurements performed by the TDR (blue dots) at a higher frequency. \label{fig:gpr_schema_all}}
\end{figure*}

We present two approaches to extend the given multi-sensor dataset and to perform a machine learning based estimation of soil moisture.
In \cref{fig:gpr_schema_all}, the GPR and TDR measurements are illustrated schematically with both simulation approaches.
The first approach originates from the \num{10} times finer temporal resolution of the TDR data in contrast to the GPR data (cf. \cref{fig:gpr_schema_meas}).
This approach generates simulated GPR data based on TDR data combined with measured GPR profiles.
The resulting data is spatially localized at the TDR positions of each plot (cf.~\cref{fig:gpr_schema_app1}).
In \cref{sec:methods:sub:app1}, we describe the first approach in detail.

The basis of the second approach are the GPR profiles along a line of each plot.
We simulate values of soil moisture (respectively TDR) along the GPR profiles by merging the measurements of TDR and GPR.
Schematically, the second approach is presented in \cref{fig:gpr_schema_app2} and described in detail in \cref{sec:methods:sub:app2}.

\subsection{Simulation of GPR data}
\label{sec:methods:sub:app1}

\begin{figure}[tb]
	\centering
	\includegraphics[width=0.4\textwidth]{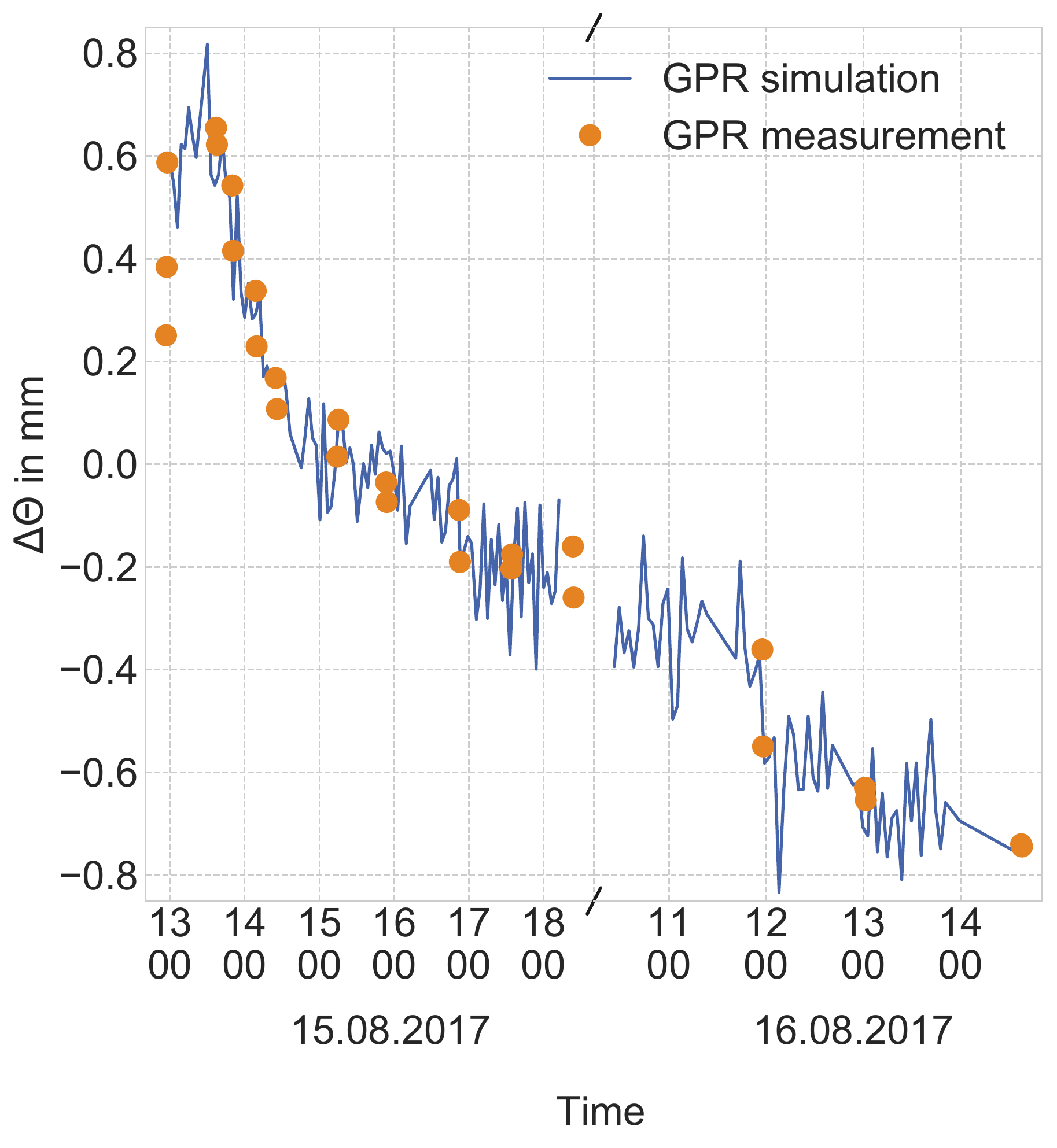}
	\caption{Exemplary GPR time series of plot 2. The GPR measurements are orange and the GPR simulation with Gaussian noise are blue. The gap in the middle represents the night between to the two measurement days.\label{fig:app1_timeseries}}
\end{figure}

For this first simulation approach, we propose two distinct simulations of the soil-moisture content as a GPR product.
First, we interpolate between different GPR measurements along the time axis (cf.~\cite{oliphant2007python}).
This interpolation ignores any TDR measurements, since they possibly bias the interpolation results. 
Additionally, we calculate Gaussian noise to these results to reduce overfitting effects.
In sum, the simulated dataset consists of \num{481} datapoints with hyperspectral and GPR data.
An exemplary interpolation is shown in \cref{fig:app1_timeseries}.

Next, we apply machine learning to link the GPR and TDR soil-moisture measurements.
We evaluate this approach with two regression models: linear regression and extremely randomized trees (ET)~\cite{geurts2006extremely}.
As before, the resulting dataset consists of \num{481} datapoints including simulated GPR data.

\subsection{Simulation of TDR data}
\label{sec:methods:sub:app2}

The second approach to extend the measured dataset is to simulate TDR soil-moisture values based on the GPR data and along the GPR profiles.
Our concept is to perform a regression with the GPR data at the TDR positions as input data and the soil-moisture values retrieved by the TDR measurements at simultaneous times as target variable.
We apply a linear interpolation (cf.~\cite{oliphant2007python}) along the (sorted) GPR measurements as well as a linear regression and an ET regression.

After the training phase, the regressor is able to estimate soil-moisture values with given GPR $\Delta\theta$ values.
As simulation, we use all GPR data except the ones located at the TDR probes. 
In sum, we rely on \num{9} of \num{10} measured $\Delta\theta$ values of each plot.
Subsequently, we match the results of the simulated soil-moisture values with the respective pixels of the hyperspectral data.
The resulting dataset consists of $825$ datapoints including simulated TDR, GPR $\Delta\theta$ values, and the hyperspectral spectrum.

\section{Results \& evaluation}
\label{sec:eval}
In the following, we evaluate the performance of the soil-moisture estimation based on hyperspectral data combined with either simulated GPR as additional input feature or simulated TDR data as target variable.
The soil-moisture regression is performed with the ET regressor due to its best performance in~\cite{keller2018is}.
We split the three generated datasets (cf. \cref{fig:gpr_schema_all}) each into a training and a test subset at a ratio of $1:1$.
As an expression of the regression performance, we present the coefficient of determination $R^2$ and the root mean squared error RMSE.

\subsection{Evaluation of the first approach}
\label{sec:eval:sub:app1}

The regression results for the first approach with the simulated GPR values as one input data are shown in \cref{tab:app1_results}.
Adding any of the simulated GPR data as additional feature to the regression model increases the $R^2$ about at least \num{20}~p.p. compared to a regression performed without this data.
The GPR data estimated by the linear regression performs the best in estimating soil moisture and achieves an  $R^2=\SI{83.3}{\percent}$.
In this specific approach, the simulated GPR data is a linear combination of several measured soil-moisture sample points.
We are aware of the TDR bias which is included in the linear and ET regression.

By contrast, the approach applying the interpolation and Gaussian noise is based exclusively on the time course of the GPR measurements.
The regression result based on the GPR interpolation is $R^2=\SI{74.4}{\percent}$.
This is a significant improvement and underlines the importance of GPR data combined with the hyperspectral data in this estimation.

\begin{table}[tb]
	\centering
	\caption{Results of the ET regression with the soil-moisture target variable and the simulated GPR data as input data. For the simulation, an interpolation, a linear regression, and an ET regression are evaluated. FI is the feature importance of the simulated GPR data in the ET regression.}
	\begin{tabular}{lSSc}
	\toprule
	\multirow{2}{*}{GPR simulation} & {$R^2$} & {RMSE} & {FI of GPR}\\
	& {in \%} &  {in \si{\milli\meter}} & { in \%}\\ 
	\midrule
	No simulated GPR 	& 53.2 & 1.09 & {-}\\
	Interpolation	 	& 74.4 & 0.81 & 31\\
	Linear regression 	& 83.3 & 0.65 & 39\\
	ET regression 	 	& 77.9 & 0.75 & 36\\
	\bottomrule
	\end{tabular}
	\label{tab:app1_results}	
\end{table}

\subsection{Evaluation of the second approach}
\label{sec:eval:sub:app2}

As before, the evaluation is performed with an ET regression on simulated sensor-like soil-moisture data as target variable with solely hyperspectral data as input features.
The results are shown in \cref{tab:app2_results}.
Based on these regression results, we conclude that the regression models struggle to precisely link the simulated soil-moisture data and the hyperspectral data.
The small number of \num{94} measured GPR datapoints combined with a poor correlation of some plots as pointed out in \cref{sec:data}, cf. \cref{fig:corr}, explain these weak regression performances.

\begin{table}[tb]
	\centering
	\caption{Results of the ET regression with the simulated soil-moisture values as target variable and the hyperspectral data as input data. For the simulation, an interpolation, a linear regression, and an ET regression are evaluated.}
	\begin{tabular}{lSS}
	\toprule
	{TDR simulation} & {$R^2$ in \%} & {RMSE in \si{\milli\meter}}\\
	\midrule
	Interpolation	 	& 21.5 & 2.16\\
	Linear regression 	& 37.5 & 0.76\\
	ET regression 	 	& 35.1 & 0.81\\
	\bottomrule
	\end{tabular}
	\label{tab:app2_results}	
\end{table}

\begin{figure}[tb]
	\centering
	\includegraphics[width=0.48\textwidth]{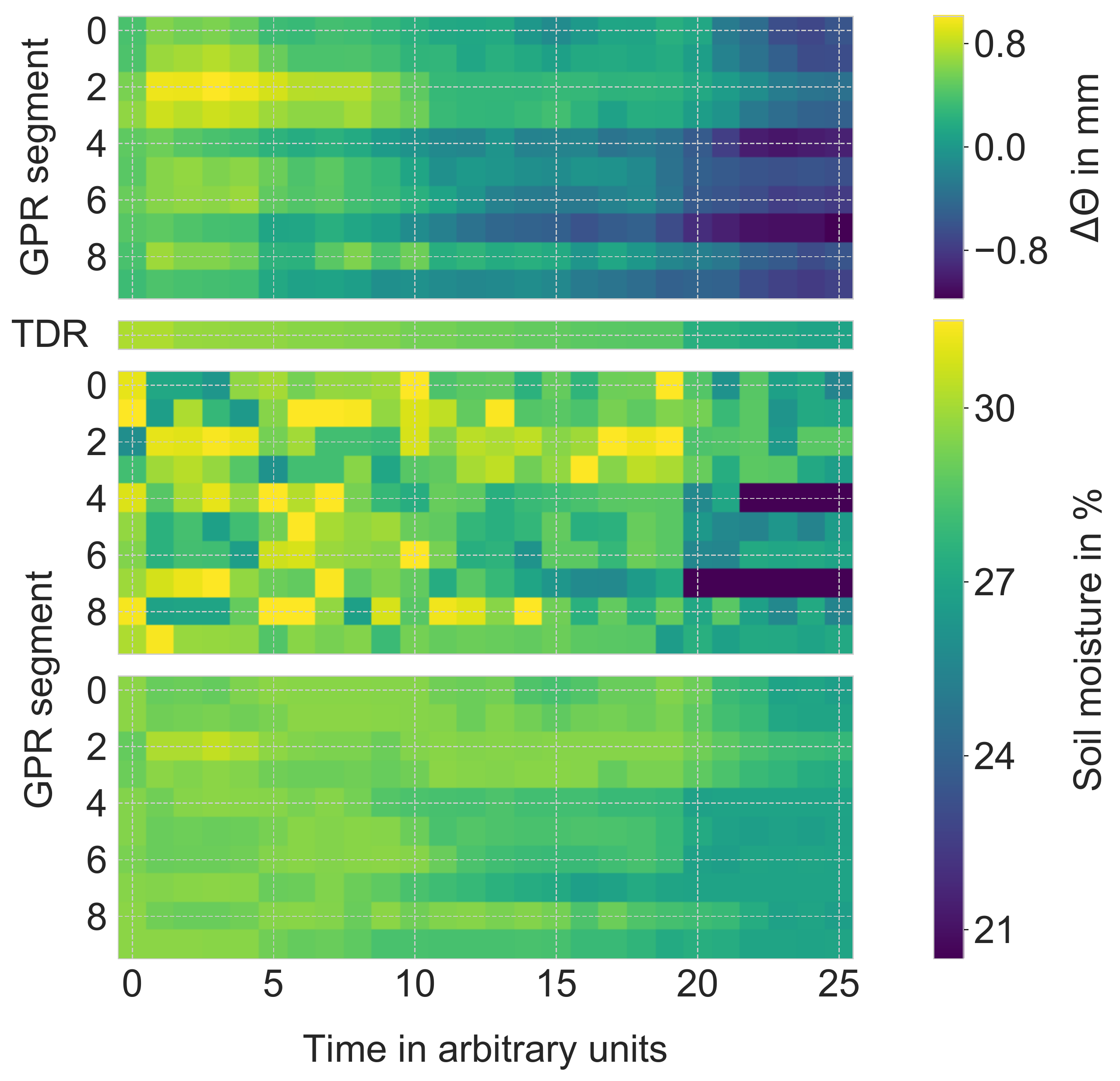}
	\caption{Distributions of the measured $\Delta\theta$ of the GPR (top), the measured TDR values (center top), the linearly interpolated soil-moisture values (center bottom), and the simulated soil-moisture values of the linear regression (bottom) of measurement plot 1.\label{fig:sim_dist}}
\end{figure}

\cref{fig:sim_dist} illustrates an exemplary soil-moisture distribution based on the interpolation and the linear regression approaches.
The distribution as a result of the latter approach is much smoother than the distribution of the interpolation.

\section{Conclusion}
\label{sec:conclusion}

In this contribution, we evaluate the potential of hyperspectral data combined with either simulated GPR or simulated TDR data to estimate soil moisture.
This estimation is performed with a machine learning model.
Based on the two different approaches, we extend the underlying sparse dataset.

In the first approach, we simulate GPR data. 
The second approach handles the simulation of soil-moisture values analog to TDR measurements.
Our results of the soil-moisture estimation reveal the potential of (simulated) GPR data as valuable input.
The simulated GPR, even as output of a common linear regression, enhances the overall performance of the soil-moisture estimation significantly. 
The fusion of simulated soil-moisture values (TDR similar) and hyperspectral data as input data for the soil-moisture estimation performs the worst.

To conclude, the soil-moisture estimation is improved by the fusion of hyperspectral and ground penetrating radar data.
In future work, we plan to extend the field campaigns with many more GPR profiles.
Thus, we are able to enhance the GPR simulation based on measured data.
The area-wide estimation of soil moisture benefits from the proposed fusion of hyperspectral and GPR data.

\section{Acknowledgements}
\label{sec:ack}

We thank Niklas Allroggen for his tedious-acquired GPR profiles.
We also thank Conrad Jackisch for the monitoring of the hydrological parameters.

\bibliographystyle{IEEEbib}
\bibliography{whispers}

\end{document}